\documentclass{article}
\usepackage{ijcai21}

\usepackage{amsmath,amsfonts,bm}

\def\eqref#1{equation~\ref{#1}}

\def\1{\bm{1}}

\DeclareMathAlphabet{\mathsfit}{\encodingdefault}{\sfdefault}{m}{sl}
\SetMathAlphabet{\mathsfit}{bold}{\encodingdefault}{\sfdefault}{bx}{n}

\newcommand{\sigmoid}{\sigma}

\DeclareMathOperator*{\argmax}{arg\,max}

\usepackage{times}
\usepackage{soul}
\usepackage{url}
\usepackage[hidelinks]{hyperref}
\usepackage[utf8]{inputenc}
\usepackage[small]{caption}
\usepackage{multirow}
\usepackage{amssymb}
\usepackage{makecell}
\usepackage{array}
\usepackage{graphicx}
\usepackage{amsmath}
\usepackage{amsthm}
\usepackage{booktabs}
\usepackage{algorithm}
\usepackage{algorithmic}

\usepackage{multirow}
\usepackage{amssymb}
\usepackage{makecell}
\usepackage{array}
\usepackage{graphicx}
\usepackage{subcaption}
\usepackage[flushleft]{threeparttable}
\usepackage{stackrel}
\usepackage{appendix}
\usepackage{hyperref}

\title{Masked Label Prediction: Unified Message Passing Model for Semi-Supervised Classification}

\author{
Yunsheng Shi\and
Zhengjie Huang\and
Shikun Feng\and
Hui Zhong\and
Wenjing Wang\and
Yu Sun
\affiliations
Baidu Inc., China
\emails
\{shiyunsheng01, huangzhengjie, fengshikun01, zhonghui03, wangwenjin02, sunyu02\}@baidu.com
}

\begin{document}
	
	\maketitle
	
	\begin{abstract}
		Graph neural network (GNN) and label propagation algorithm (LPA) are both message passing algorithms, which have achieved superior performance in semi-supervised classification. GNN performs \emph{feature propagation} by a neural network to make predictions, while LPA uses \emph{label propagation} across graph adjacency matrix to get results. However, there is still no effective way to directly combine these two kinds of algorithms. To address this issue, we propose a novel {\bf Uni}fied {\bf M}essage {\bf P}assaging Model (UniMP) that can incorporate \emph{feature} and \emph{label propagation} at both training and inference time. First, UniMP adopts a Graph Transformer network, taking feature embedding and label embedding as input information for propagation. Second, to train the network without overfitting in self-loop input label information, UniMP introduces a masked label prediction strategy, in which some percentage of input label information are masked at random, and then predicted. UniMP conceptually unifies feature propagation and label propagation and is empirically powerful. It obtains new state-of-the-art semi-supervised classification results in Open Graph Benchmark (OGB). 
	\end{abstract}
	
	\section{Introduction}
	There are various scenarios in the world, e.g., recommending related news, discovering new drugs, or predicting social relations, which can be described as graph structures. And many methods have been proposed to optimize these graph-based problems and achieved significant success in many related domains such as predicting nodes' properties ~\cite{yang2016revisiting,kipf2016semi}, relation linking ~\cite{grover2016node2vec,battaglia2018relational}, and graph classification ~\cite{duvenaud2015convolutional,niepert2016learning}.
	
	In the task of semi-supervised node classification, we are required to learn with labeled examples and then make predictions for those unlabeled ones. To better classify the nodes' labels in the graph, based on the Laplacian smoothing assumption~\cite{li2018deeper,xu2018representation}, the message passing models were proposed to aggregate the information from its connected neighbors in the graph, acquiring
	enough facts to produce a more robust prediction for unlabeled nodes. Generally, there are two main kinds of methods to implement message passing model, Graph Neural Networks (GNNs) ~\cite{kipf2016semi,hamilton2017inductive,xu2018representation,liao2019lanczosnet,xu2018powerful} and Label Propagation Algorithms (LPAs) ~\cite{zhu2003semi,zhang2007hyperparameter,wang2007label,karasuyama2013manifold,gong2016label,liu2019learning}. GNNs combine graph structures by propagating and aggregating node features through several neural layers, getting predictions from \emph{feature propagation}. While LPAs make predictions for unlabeled instances by \emph{label propagation} iteratively.
	
	Since GNN and LPA are based on the same assumption, making semi-supervised classifications by information propagation, there is an intuition that incorporating them together for boosting performance. Some superior studies have proposed their graph models based on it. For example, APPNP ~\cite{klicpera2019predict} and TPN~\cite{liu2019learning} using GNN predict soft labels and then propagate them, and GCN-LPA~\cite{wang2019unifying} uses LPA to regularize their GNN model. However, as shown in Table \ref{table:comparision}, aforementioned methods
	still can not directly incorporate GNN and LPA within a message passing model, \emph{propagating} \emph{feature} and \emph{label} in both training and inference procedure.
	\begin{table}[tbp]
		\setlength{\abovecaptionskip}{0.cm}
		\setlength{\belowcaptionskip}{-.2cm}
		\begin{center}
			\resizebox{0.91\columnwidth}{!}{%
				\begin{tabular}{cccccc}
					\hline
					\multicolumn{1}{c}{ }  &\multicolumn{2}{c}{ Training} &\multicolumn{2}{c}{ Inference} \\
					\multicolumn{1}{c}{\bf Model}  &\multicolumn{1}{c}{\bf Feature} &\multicolumn{1}{c}{\bf Label} &\multicolumn{1}{c}{\bf Feature} &\multicolumn{1}{c}{\bf Label}	\\
					\hline 
					LPA         & &\makecell[c]{\checkmark} & &\makecell[c]{\checkmark}   \\
					GCN             &\makecell[c]{\checkmark}  &&\makecell[c]{\checkmark}  & \\
					APPNP     &\makecell[c]{\checkmark}    &   &\makecell[c]{\checkmark}    & \\
					GCN-LPA      &\makecell[c]{\checkmark}    & \makecell[c]{\checkmark}    &\makecell[c]{\checkmark}    & \\
					\bf UniMP (Ours)     &\makecell[c]{\checkmark}    & \makecell[c]{\checkmark}    &\makecell[c]{\checkmark}    &\makecell[c]{\checkmark}  \\
					\hline
				\end{tabular}
			}
		\end{center}
		
		\caption{Comparison the input information that message passing models use in training and inference.}
		\vspace{-0.32cm}
		\label{table:comparision}
	\end{table}
	
	
	In this work, we propose a {\bf Uni}fied {\bf M}essage {\bf P}assing model (UniMP) to address the aforementioned issue with two simple but effective ideas: (a) combing node features propagation with labels and (b) masked label prediction. Previous GNN-based methods only take node features as input with the partial observed node labels for supervised training. And they discard the observed labels during inference. UniMP utilizes both node features and labels in both training and inference stages. It uses the embedding technique to transform the partial node labels from one-hot to dense vector likes node features. And a multi-layer Graph Transformer network takes them as input to perform attentive information propagation between nodes. Therefore, each node can aggregate both features and labels information from its neighbors. Since we have taken the node label as input, using it for supervised training will cause the label leakage problem. The model will overfit in the self-loop input label while performing poor in inference. To address this issue, we propose a masked label prediction strategy, which randomly masks some training instances' label and then predicts them to overcome label leakage. This simple and effective training method is drawn the lesson from masked word prediction in BERT~\cite{devlin2018bert}, and simulates the procedure of transducing labels information from labeled to unlabeled examples in the graph. 
	
	We evaluate our UniMP model on three semi-supervised classification datasets in the Open Graph Benchmark (OGB), where our new methods achieve the new state-of-the-art results in all tasks, gaining 82.56\% ACC in \emph{ogbn-products}, 86.42\% ROC-AUC in \emph{ogbn-proteins} and 73.11\% ACC in \emph{ogbn-arxiv}. We also conduct ablation studies for our UniMP model, to evaluate the effectiveness of our unified method. Besides, we make the most thorough analysis of how the label propagation boosts our model's performance.
	
	\section{Preliminaries}
	In this section, we briefly review the related work and along the way, introduce our notation. We denote a graph as $G = (V,E)$, where $V$ denotes the nodes in the graph with $|V| = n$ and $E$ denotes edges. The nodes are described by the feature matrix $X \in \mathbb{R} ^{n \times m}$, which usually are dense vectors with $m$ dimension, and the target class matrix $Y \in \mathbb{R} ^{n \times c}$, with the number of classes $c$. The adjacency matrix $A = [a_{i,j}] \in \mathbb{R}^{n \times n}$ is used to describe graph $G$, and the diagonal degree matrix is denoted by $D = $ diag$(d_1,d_2,...,d_n)$ , where $d_i = \sum_{j} a_{ij}$ is the degree of node $i$. A normalized adjacency matrix is defined as $D^{-1}A$ or $D^{- \frac{1}{2}}AD^{- \frac{1}{2}}$, and we adopt the first definition in this paper.
	
	\paragraph{Graph Neural Networks.} In semi-supervised node classification, GCN \cite{kipf2016semi} is one of the most classical models based on the Laplacian smoothing assumption. GCN transforms and propagates node features $X$ across the graph by several layers, including linear layers and nonlinear activation to build the approximation
	of the mapping: $X \rightarrow Y$. The feature propagation scheme of GCN in layer $l$ is:
	\begin{equation}
	\begin{split}
	H^{(l+1)}&=\sigma(D^{-1}AH^{(l)}W^{(l)})\\
	Y&=f_{out}(H^{(L)})
	\end{split}
	\label{equ:1}
	\end{equation}
	where the $\sigma$ is an activation function, $W^{(l)}$ is the trainable weight in the $l$-th layer, and the $H^{(l)}$ is the $l$-th layer representations of nodes. $H^{(0)}$ is equal to node input features $X$. Finally, a $f_{out}$ output layer is applied on the final representation to make prediction for $Y$.
	
	\paragraph{Label propagation algorithms.} Traditional algorithms like Label Propagation Algorithm (LPA) only utilizes labels and relations between nodes to make prediction. LPA assumes the labels between connected nodes are similar and propagates the labels iteratively across the graph. Given an initial label matrix $\hat{Y}^{(0)} $, which consists of one-hot label indicator vectors $\hat{y}_i^{0}$ for the labeled nodes or zeros vectors for the unlabeled. A simple iteration equation of LPA is formulated as following:
	\begin{equation}
	\begin{split}
	\hat{Y}^{(l+1)}&=D^{-1}A\hat{Y}^{(l)}
	\end{split}
	\end{equation}
	Labels are propagated from each other nodes through a normalized adjacency matrix $D^{-1}A$. 
	
	\paragraph{Combining GNN and LPA.} Recently, there is a trend to combine GNN and LPA in semi-classification tasks in the community. APPNP~\cite{klicpera2019predict} and TPN~\cite{liu2019learning} propose to use GCN to predict soft labels and then propagate them with Personalized Pagerank. However, these works still only considered the partial node labels as the supervision training signal. GCN-LPA is most relevant to our work, as they also take the partial node labels as input. However, they combine the GNN and LPA in a more indirect way, only using the LPA in training to regularize the weight edges of their GAT model. While our UniMP directly combines GNN and LPA within a network, propagates the node features and labels 
in both training and predicting. Moreover, unlike GCN-LPA whose regularization strategy can only be used in those GNNs with trainable weight edge such as GAT~\cite{velivckovic2017graph}, GAAN~\cite{zhang2018gaan}, our training strategy can be easily extended in kinds of GNNs such as GCN and GAT to further improve their performance. We will describe our approach more specifically in the next section.
	
	\section{Unified Message Passing Model}
	\begin{figure}[tbp]
		\centering
		\includegraphics[width=\columnwidth,trim=1cm 2cm 2cm 0.2cm, clip]{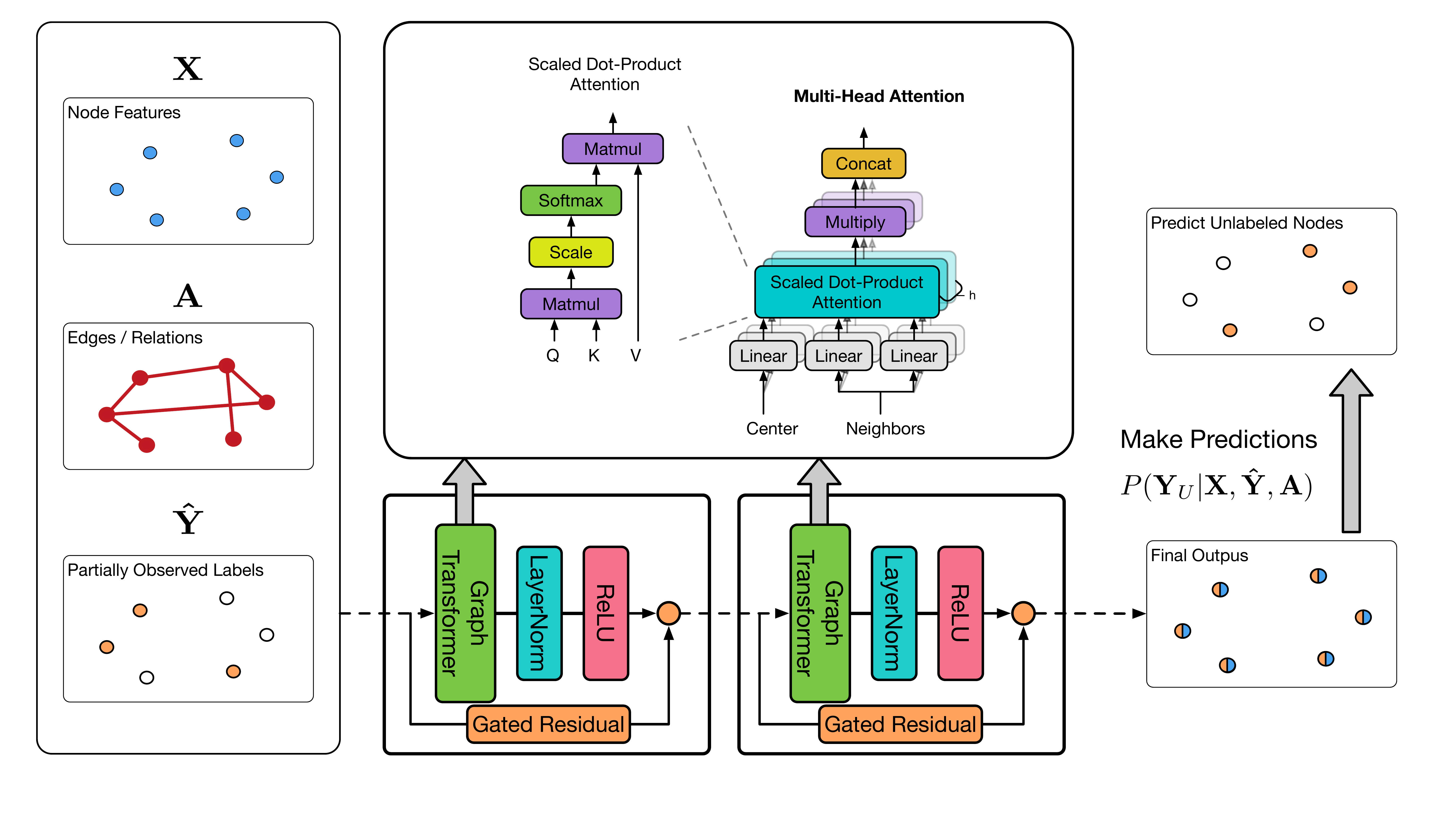}
		\caption{
			The architecture of UniMP.
		}
		\vspace{-0.25cm}
		\label{figure:unimp_architecture}
	\end{figure}
	
	As shown in Figure \ref{figure:unimp_architecture}, given the node feature $X$ and partial observed labels $\hat{Y}$, we employ a Graph Transformer, jointly using label embedding to combine the aforementioned feature and label propagation together, constructing our UniMP model. Moreover, a masked label prediction strategy is introduced to train our model to prevent label leakage problem. 
	
	\subsection{Graph Transformer}
	
	Since Transformer \cite{vaswani2017attention,devlin2018bert} has been proved being powerful in NLP, we adopt its vanilla multi-head attention into graph learning with taking into account the case of edge features. Specifically, given node features $H^{(l)}=\{h^{(l)}_1,h^{(l)}_2,...,h^{(l)}_n \}$, we calculate multi-head attention for each edge from $j$ to $i$ as following:
	\begin{equation}
	\begin{split}
	q_{c,i}^{(l)} &=W_{c,q}^{(l)} h^{(l)}_{i}+b^{(l)}_{c,q} \\
	k_{c,j}^{(l)} &=W_{c,k}^{(l)} h^{(l)}_{j}+b^{(l)}_{c,k} \\
	e_{c,ij} &=W_{c,e} e_{ij}+b_{c,e} \\
	\alpha_{c,ij}^{(l)} &= \frac{ \langle q_{c,i}^{(l)} , k_{c,j}^{(l)} +e_{c, ij}  \rangle }{\sum_{u \in \mathcal{N}(i)} \langle q_{c,i}^{(l)} , k_{c,u}^{(l)} + e_{c, iu} \rangle }
	\end{split}
	\label{eq:3}
	\end{equation}
	where $\langle q,k \rangle =\exp(\frac{q^{T}k}{\sqrt{d}} )$ is exponential scale dot-product function and $d$ is the hidden size of each head.
	For the $c$-th head attention, we firstly transform the source feature $h^{(l)}_i$ and distant feature $h^{(l)}_j$ into query vector $q^{(l)}_{c,i} \in \mathbb{R}^{d}$and key vector $k^{(l)}_{c,j} \in \mathbb{R}^{d}$ respectively using different trainable parameters $W_{c,q}^{(l)} $, $W_{c,k}^{(l)} $, $b^{(l)}_{c,q}$, $b^{(l)}_{c,k}$. The provided edge features $e_{ij}$ will be encoded and added into key vector as additional information for each layer.
	
	After getting the graph multi-head attention, we make a message aggregation from the distant $j$ to the source $i$:
	\begin{equation}
	\begin{aligned}
	v_{c,j}^{(l)} &=W_{c,v}^{(l)} h^{(l)}_{j}+b^{(l)}_{c,v} \\
	\hat{h}_{i}^{(l+1)} &=  \Big\Vert_{c=1}^{C}  \big[ \sum_{j \in \mathcal{N}(i)} \alpha^{(l)}_{c,ij}(v^{(l)}_{c,j} + e_{c,ij}) \big] \\
	\end{aligned}
	\end{equation}
	where the $\parallel$ is the concatenation operation for $C$ head attention. Comparing with the Equation \ref{equ:1}, multi-head attention  matrix replaces the original normalized adjacency matrix as transition matrix for message passing. The distant feature $h_j$ is transformed to $v_{c,j} \in \mathbb{R}^{d}$ for weighted sum. 
	
	In addition, inspired by Li~\shortcite{li2019deepgcns} and Chen~\shortcite{chen2020simple}, we propose to use a gated residual connection between layers as shown in Equation \ref{eq:4} to prevent our model from over-smoothing.
	\vspace{-1mm}
	\begin{equation}
	\resizebox{.91\linewidth}{!}{$
		\begin{aligned}
		r_{i}^{(l)} &=W_{r}^{(l)} h^{(l)}_{i}+b^{(l)}_{r} \\
		\beta^{(l)}_{i} &= \text{sigmoid}(W_{g}^{(l)}[\hat{h}^{(l+1)}_{i}; r^{(l)}_i; \hat{h}^{(l+1)}_{i} -  r^{(l)}_i]) \\
		h_{i}^{(l+1)} &= \text{ReLU}(\text{LayerNorm}((1- \beta_{i}^{(l)} )\hat{h}_{i}^{(l+1)}+\beta_{i}^{(l)}  r_{i}^{(l)} )) \\
		\end{aligned}$}
	\label{eq:4}
	\end{equation}
	
	Specially, similar to GAT, if we apply the Graph Transformer on the last output layer, we will employ averaging for multi-head output and remove the non-linear transformation as following: 
	\vspace{-1mm}
	\begin{equation}
	\begin{split}
	\hat{h}_{i}^{(l+1)} &=  \frac{1}{C}\sum_{c=1}^{C}  \big[ \sum_{j \in \mathcal{N}(i)} \alpha^{(l)}_{c,ij}(v^{(l)}_{c,j} + e_{c,ij}^{(l)}) \big] \\
	h_{i}^{(l+1)} &= (1- \beta_{i}^{(l)} )\hat{h}_{i}^{(l+1)}+\beta_{i}^{(l)}  r_{i}^{(l)}\\
	\end{split}
	\end{equation}
	
	\subsection{Label Embedding and Propagation}
	We propose to embed the partially observed labels into the same space as node features: $ \hat{Y}\in\mathbb{R}^{n\times c} \rightarrow \hat{Y}_d \in \mathbb{R}^{n\times m}$, which consist of the label embedding vector for labeled nodes and zeros vectors for the unlabeled. And then, we combine the label propagation into Graph Transformer by simply adding the node features and labels vectors together as propagation information $(H^{0}=X+\hat{Y}_d) \in \mathbb{R}^{n \times m}$. We can prove that by mapping partially-labeled $\hat{Y}$ and node features $X$ into the same space and adding them up, our model is unifying both label propagation and feature propagation within a shared message passing framework. Let's take $\hat{Y}_d=\hat{Y}W_{d}$ and $A^{*}$ to be normalized adjacency matrix $D^{-1}A$ or the attention matrix from our Graph Transformer likes Equation \ref{eq:3}. Then we can find that:
	\vspace{-1mm}
	\begin{equation}
	\begin{split}
	H^{(0)} &= X + \hat{Y}W_{d} \\
	H^{(l+1)} &= \sigmoid(((1 - \beta) A^{*} + \beta I ) H^{(l)}W^{(l)})
	\end{split}
	\end{equation}
	where $\beta$ can be the gated function like Equation \ref{eq:4} or a pre-defined hyper-parameters like APPNP \cite{klicpera2019predict}. For simplification, we let $\sigmoid$ function as identity function, then we can get:
	\vspace{-1mm}
	\begin{equation}
	\resizebox{.91\linewidth}{!}{$
		\begin{aligned}
		H^{(l)} =& ((1 - \beta) A^{*} + \beta I )^{l} (X + \hat{Y}W_{d}) W^{(1)} W^{(2)}\dots W^{(l)} \\
		=& ((1 - \beta) A^{*} + \beta I )^{l}X W  + ((1 - \beta) A^{*} + \beta I )^{l}\hat{Y}W_{d} W\\
		\end{aligned}$}
	\end{equation}
	where $W=W^{(1)} W^{(2)}\dots W^{(l)}$. Then we can find that our model can be approximately decomposed into feature propagation $ ((1 - \beta) A^{*} + \beta I )^{l}X W $ and label propagation $ ((1 - \beta) A^{*} + \beta I )^{l}\hat{Y}W_{d} W$. 
	
	\subsection{Masked Label Prediction}
	
	Previous works on GNNs seldom consider using the partially observed labels $\hat{Y}$ in both training and inference stages. They only take those labels information as ground truth target to supervised train their model's parameters $\theta$ with given $X$ and $A$:
	\vspace{-1.5mm}
	\begin{equation} 
	\label{equ:ind_formulate}
	\argmax_{\theta} \quad \log p_{\theta}(\hat{Y}| X, A) = \sum_{i=1}^{\hat{V}} \log p_{\theta}(\hat{y}_i | X, A)\\
	\end{equation}
	where $\hat{V}$ represents the partial nodes with labels. However, our UniMP model propagates node features and labels to make prediction: $p(y | X, \hat{Y}, A)$. Simply using above objective for our model will make the label leakage in the training stage, causing poor performance in inference. Learning from BERT, which masks input words and makes predictions for them to pretrain their model (masked word prediction), we propose a masked label prediction strategy to train our model. During training, at each step, we corrupt the $\hat{Y}$ into $\tilde{Y}$ by randomly masking a portion of node labels to zeros and keep the others remain, which is controlled by a hyper-parameter called label\_rate. Let those masked labels be $\bar{Y}$, our objective function is to predict $\bar{Y}$ with given $X$, $\tilde{Y}$ and $A$:
	\vspace{-1.5mm}

	\begin{equation} 
	\label{equ:ind_formulate_v2}
	\argmax_{\theta} \text{\ } \log p_{\theta}(\bar{Y}| X,\tilde{Y}, A) = \sum_{i=1}^{\bar{V}} \log p_{\theta}(\bar{y}_i | X, \tilde{Y}, A)
	\end{equation}
	where $\bar{V}$ represents those nodes with masked labels. In this way, we can train our model without the leakage of self-loop labels information. And during inference, we will employ all $\hat{Y}$ as input labels to predict the remaining unlabeled nodes.
	
	\section{Experiments}

We propose a Unified Message Passing Model (UniMP) for semi-supervised node classification, which incorporates the feature and label propagation jointly by a Graph Transformer and employs a masked label prediction strategy to optimize it. We conduct the experiments on the Node Property Prediction of Open Graph Benchmark (OGBN), which includes several various challenging and large-scale datasets for semi-supervised classification, split in the procedure that closely matches the real-world application \cite{hu2020open}. To verify our models effectiveness, we compare our model with others state-of-the-art (SOTA) models in \emph{ogbn-products}, \emph{ogbn-proteins} and \emph{ogbn-arxiv} three OGBN datasets. We also provide more experiments and comprehensive ablation studies to show our motivation more intuitively, and how LPA improves our model to achieve better results.

\subsection{Datasets and Experimental Settings}

\begin{table}[htbp]
	\setlength{\abovecaptionskip}{-0cm}
	\begin{center}
		\setlength{\tabcolsep}{1.2mm}{
		\resizebox{\columnwidth}{!}{%
			\begin{tabular}{l|ccccc}
			\hline
				{\bf Name}  & {\bf Node} &  {\bf Edges}  &  {\bf Tasks}    &  {\bf Task Type} &  {\bf Metric}	\\
				\hline 
				ogbn-products  &2,449,029&61,859,140 & 1 &Multi-class class &Accuracy \\
				ogbn-proteins   &132,534&39,561,252&112  &Binary class  &ROC-AUC\\
				ogbn-arxiv     &169,343&1,166,243&1 &Multi-class class  &Accuracy \\
				\hline
		\end{tabular}}}
	\end{center}
	\caption{Dataset statistics of OGB node property prediction}
	\label{table:ogb}
	\vspace{-4mm}
\end{table}

\paragraph{Datasets.} Most of the frequently-used graph datasets are extremely small compared to graphs found in real applications. And the performance of GNNs on these datasets is often unstable due to several issues including
their small-scale nature, non-negligible duplication or leakage rates, unrealistic data splits \cite{hu2020open}. Consequently, we conduct our experiments on the recently released datasets of Open Graph Benchmark (OGB) \cite{hu2020open}, which overcome the main drawbacks of commonly used datasets and thus
are much more realistic and challenging. OGB datasets cover a variety of real-world applications and span several important domains ranging from social and information networks to biological networks, molecular graphs, and knowledge graphs. They also span a variety of prediction tasks at the level of nodes, graphs, and links/edges. As shown in table \ref{table:ogb}, in this work, we performed our experiments on the three OGBN datasets with different sizes and tasks for getting credible result, including \emph{ogbn-products} about 47 products categories classification with given 100-dimensional nodes features, \emph{ogbn-proteins} about 112 kinds of proteins function prediction with given 8-dimensional edges features and \emph{ogbn-arxiv} about 40-class topics classification with given 128 dimension nodes features. More details about these datasets are provided in appendix A in the supplementary file.


\begin{table}[htbp]
	\setlength{\abovecaptionskip}{-0cm}
	
	\begin{center}
	    \resizebox{\columnwidth}{!}{%
		\begin{tabular}{cccc}
		\hline
			& {\bf ogbn-products} &  {\bf ogbn-proteins	}  &  {\bf ogbn-arxiv} \\
			\hline
			sampling\_method &NeighborSampling & Random Partition & Full-batch \\
			num\_layers &  3 & 7 & 3\\ 
			hidden\_size & 128 & 64 & 128\\ 
			num\_heads & 4 & 4 & 2\\ 
			dropout  & 0.3 & 0.1 & 0.3\\ 
			lr & 0.001 & 0.001 & 0.001\\ 
			weight\_decay& * & * & 0.0005\\ 
			label\_rate& 0.625 & 0.5 & 0.625\\ 
			\hline
		\end{tabular}
		}
	\end{center}
	\caption{The hyper-paramerter setting of our model}
	\label{table:params}
	\vspace{-4mm}
\end{table}

\paragraph{Implementation details.}  As mentioned above, these datasets are different from each other in sizes or tasks. So we evaluate our model on them with different sampling methods following previous studies~\cite{li2020deepergcn}, getting credible comparison results. In \emph{ogbn-products} dataset, we use NeighborSampling with size =10 for each layer to sample the subgraph during training and use full-batch for inference. In \emph{ogbn-proteins} dataset, we use Random Partition to split the dense graph into subgraph to train and test our model. As for small-size \emph{ogbn-arxiv} dataset, we just apply full batch for both training and test. We set the hyper-parameter of our model for each dataset in Table \ref{table:params}, and the label\_rate means the percentage of labels we preserve during applying masked label prediction strategy. We use Adam optimizer with lr = 0.001 to train our model. Specially, we set weight decay to 0.0005 for our model in small-size \emph{ogbn-arxiv} dataset to prevent overfitting. More details about the tuned hyper-parameters are provided in appendix B in the supplementary file.
 
 
 
\subsection{Comparison with SOTA Models}
Baseline and other comparative SOTA models are provided by OGB leaderboard. And all these results are guaranteed to be reproducible with open source codes. Following the requirement of OGB, we run our experimental results for each dataset 10 times and report the mean and standard deviation. As shown in Table \ref{table:ret_products}, Table \ref{table:ret_proteins}, and Table \ref{table:ret_arxiv}, our unified model outperform all other comparative models in three OGBN datasets.
Since most of the compared models only consider optimizing their models for the features propagation, these results demonstrate that incorporating label propagation into GNN models can bring significant improvements.
Specifically, we gain 82.56\% ACC in \emph{ogbn-products}, 86.42\% ROC-AUC in \emph{ogbn-proteins}, which achieves about 0.6-1.6\%  absolute improvements compared to the newly SOTA methods like DeeperGCN \cite{li2020deepergcn}. In \emph{ogbn-arxiv}, our method gains 73.11\% ACC, achieve 0.37\% absolute improvements compared to GCNII \cite{chen2020simple}, whose parameters are four times larger than ours. 

%
\begin{table}[htbp]
	\setlength{\abovecaptionskip}{0.cm}
	\setlength{\belowcaptionskip}{-.2cm}
	\begin{center}
	    \resizebox{\columnwidth}{!}{%
		\begin{tabular}{l|cccc}
		    \hline
			{\bf Model}  & {\bf Test Accuracy	} &  {\bf Validation Accuracy	}  &  {\bf Params} \\
			\hline 
			GCN-Cluster \cite{chiang2019cluster} & 0.7897 $\pm$ 0.0036 & 0.9212 $\pm$ 0.0009 & 206,895 \\
			GAT-Cluster & 0.7923 $\pm$ 0.0078 & 0.8985 $\pm$ 0.0022& 1,540,848 \\ 
			GAT-NeighborSampling & 0.7945 $\pm$ 0.0059 & - & 1,751,574 \\
			GraphSAINT \cite{zeng2019graphsaint} & 0.8027 $\pm$ 0.0026 & - & 331,661 \\
			DeeperGCN  \cite{li2020deepergcn} & 0.8090 $\pm$ 0.0020 & 0.9238 $\pm$ 0.0009 & 253,743  \\
			\hline 
			{\bf UniMP} &  \bf{0.8256} $\pm$ \bf{0.0031} & \bf{ 0.9308} $\pm$ \bf{0.0017} & 1,475,605 \\
			\hline
		\end{tabular}}
	\end{center}
	\caption{Results for ogbn-products}
	\label{table:ret_products}
	\vspace{-4mm}
\end{table}

\begin{table}[htbp]
	\setlength{\abovecaptionskip}{0.cm}
	\setlength{\belowcaptionskip}{-.2cm}
	\begin{center}
     	\resizebox{\columnwidth}{!}{%
		\begin{tabular}{l|cccc}
		    \hline 
			{\bf Model}  & {\bf Test ROC-AUC} &  {\bf Validation ROC-AUC	}  &  {\bf Params} \\
			\hline 
			GaAN \cite{zhang2018gaan} & 0.7803 $\pm$ 0.0073 & - & - \\
			GeniePath-BS \cite{liu2020bandit} & 0.7825 $\pm$ 0.0035& - & 316,754 \\ 
			MWE-DGCN& 0.8436 $\pm$ 0.0065 & 0.8973$\pm$ 0.0057&538,544 \\
			DeepGCN \cite{li2019deepgcns}& 0.8496 $\pm$ 0.0028 & 0.8921 $\pm$ 0.0011 & 2,374,456 \\
			DeeperGCN \cite{li2020deepergcn}  & 0.8580 $\pm$ 0.0017 & 0.9106 $\pm$ 0.0016 & 2,374,568  \\
			\hline 
			{\bf UniMP} & \bf{0.8642} $\pm$\bf{0.0008} & \bf{0.9175} $\pm$ \bf{0.0007} &1,909,104\\
			\hline 
		\end{tabular}}
	\end{center}
	\caption{Results for ogbn-proteins}
	\label{table:ret_proteins}
	\vspace{-4mm}
\end{table}

\begin{table}[htbp]
	\setlength{\abovecaptionskip}{0.cm}
	\setlength{\belowcaptionskip}{-.2cm}
	\begin{center}
	    \resizebox{\columnwidth}{!}{%
		\begin{tabular}{l|cccc}
		    \hline 
			{\bf Model}  & {\bf Test Accuracy	} & {\bf Validation Accuracy}  & {\bf Param} \\
			\hline 
			DeeperGCN \cite{li2020deepergcn} & 0.7192 $\pm$ 0.0016 & 0.7262 $\pm$ 0.0014 &  1,471,506 \\ 
			GaAN \cite{zhang2018gaan} & 0.7197 $\pm$ 0.0024 & - &  1,471,506 \\ 
			DAGNN \cite{liu2020towards} & 0.7209 $\pm$ 0.0025 & -  & 1,751,574 \\
			JKNet \cite{xu2018representation} & 0.7219 $\pm$ 0.0021 & 0.7335 $\pm$ 0.0007 &  331,661 \\
			GCNII \cite{chen2020simple}  & 0.7274 $\pm$ 0.0016  & - & 2,148,648  \\
 			\hline 
			{\bf UniMP} & \bf{0.7311} $\pm$ \bf{0.0021} & \bf{0.7450} $\pm$ \bf{0.0005} &473,489 \\
			\hline 
		\end{tabular}}
	\end{center}
	\caption{Results for ogbn-arxiv}
	\label{table:ret_arxiv}
	\vspace{-4mm}
\end{table}

\begin{figure*}[htbp]
\setlength{\belowcaptionskip}{-.2cm}
	\centering
	\begin{minipage}[c]{0.03\textwidth}
	\subcaption{}
	\label{figure:fig1}
	\end{minipage}
	\begin{minipage}[c]{0.20\textwidth}
	    \includegraphics[width=\linewidth]{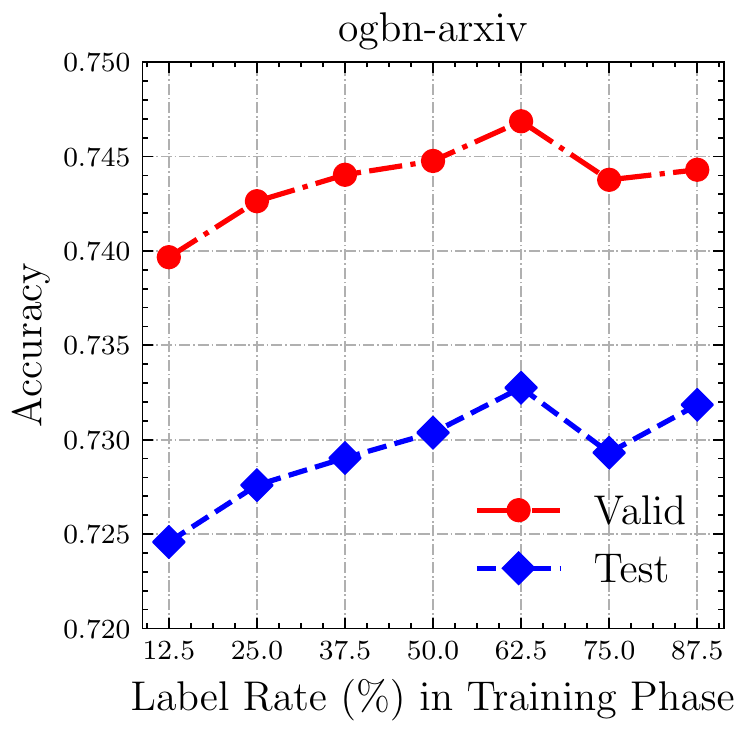}
	\end{minipage}
    \begin{minipage}[c]{0.03\textwidth}
    \subcaption{}
	\label{figure:fig2}
	\end{minipage}
	\begin{minipage}[c]{0.20\textwidth}
		\includegraphics[width=\linewidth]{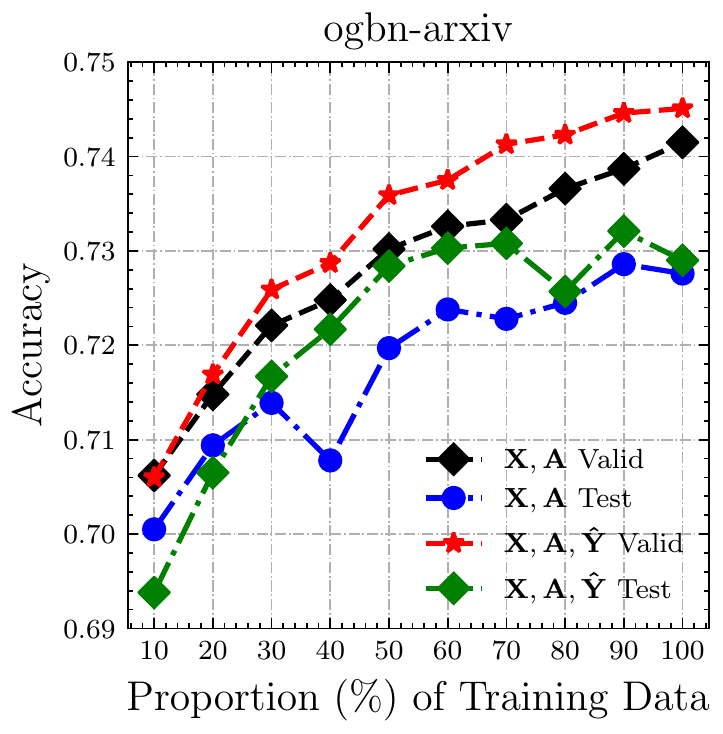}
    \end{minipage}
    \begin{minipage}[c]{0.03\textwidth}
    \subcaption{}
	\label{figure:fig3}
	\end{minipage}
	\begin{minipage}[c]{0.20\textwidth}
		\includegraphics[width=\linewidth]{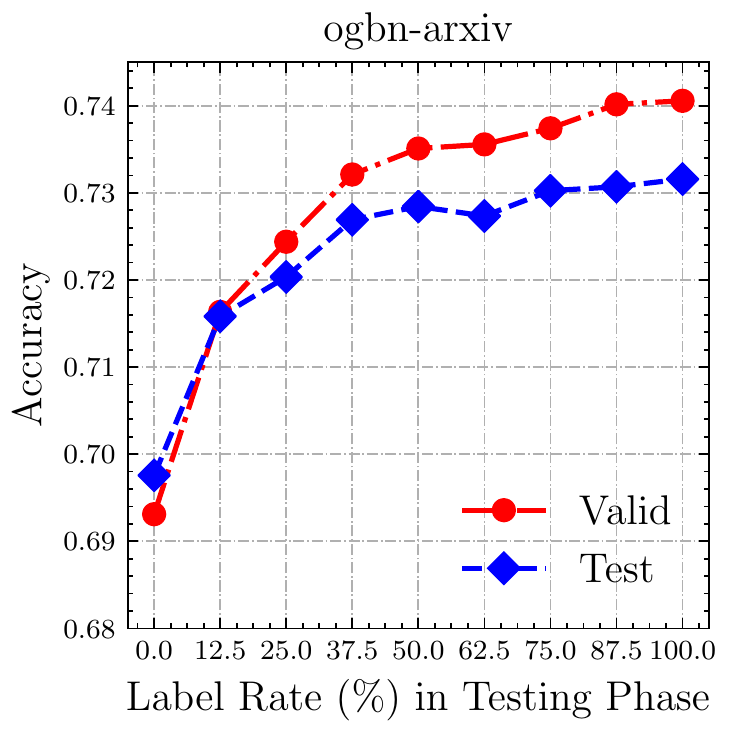}
	\end{minipage}
	\begin{minipage}[c]{0.03\textwidth}
    \subcaption{}
	\label{figure:fig4}
	\end{minipage}
	\begin{minipage}[c]{0.20\textwidth}
		\includegraphics[width=\linewidth]{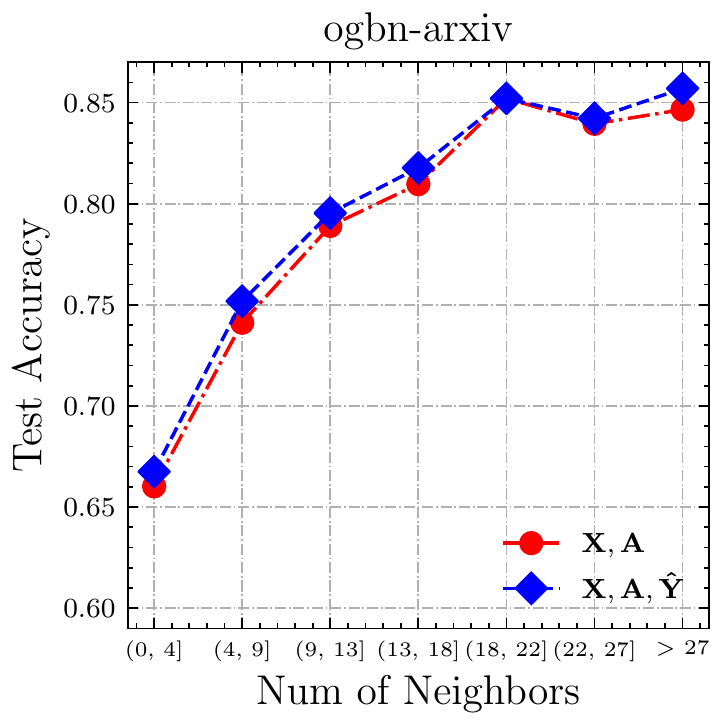}
	\end{minipage}
	\caption{Exploration of how label coverage affects label propagation: (a) Training with different label\_rate; (b) Training with different proportion of labeled data; (c) Testing with different label\_rate; (d) Test accuracy with different neighbors. }
		\label{figure:fig_curves}
	
\end{figure*}

\begin{table*}[htbp]
	\centering
   
	\small
	\setlength{\tabcolsep}{5mm}{
	\begin{tabular}{c|c|c|c|c}
		\hline
		&        & \multicolumn{3}{c}{\textbf{Datasets}}                      \\ \cline{3-5} 
		\multirow{-2}{*}{\textbf{Inputs}}        & \multirow{-2}{*}{\textbf{Model}} & \multicolumn{1}{c|}{\textbf{\begin{tabular}[c]{@{}c@{}}ogbn-products   \\ Test ACC\end{tabular}}} & \multicolumn{1}{c|}{\textbf{\begin{tabular}[c]{@{}c@{}}ogbn-proteins    \\  Test ROC-AUC\end{tabular}}} & \multicolumn{1}{c}{\textbf{\begin{tabular}[c]{@{}c@{}}ogbn-arxiv \\ Test ACC\end{tabular}}} \\ \hline
		$\mathbf{X}$   & Multilayer Perceptron    &       0.6106 $\pm$ 0.0008                    &          0.7204 $\pm$ 0.0048    &               0.5765 $\pm$ 0.0012 \\ \hline
		& GCN    &    0.7851 $\pm$ 0.0011         &           0.8265 $\pm$ 0.0008            &   0.7218 $\pm$ 0.0014                  \\ 
		& GAT    &     0.8002 $\pm$ 0.0063                &    0.8376 $\pm$ 0.0007                   &     0.7246 $\pm$ 0.0013              \\ 
		\multirow{-3}{*}{$\mathbf{X,A}$}         & Graph Transformer                      &   0.8137 $\pm$ 0.0047                  &           0.8347  $\pm$  0.0014           &       0.7292 $\pm$ 0.0010             \\ \hline
		& GCN    &          0.7832 $\pm$ 0.0013           &   0.8083 $\pm$ 0.0021                   &                      0.7018  $\pm$ 0.0009        \\ 
		& GAT    &               0.7751 $\pm$ 0.0054      &  0.8247 $\pm$ 0.0033                    &       0.7055 $\pm$ 0.0012            \\
		\multirow{-3}{*}{$\mathbf{A,\hat{Y}}$}   & Graph Transformer &          0.7987 $\pm$ 0.0104           &                  0.8160 $\pm$ 0.0007    &       0.7090 $\pm$ 0.0007 \\ \hline
		& GCN    &            0.7987 $\pm$ 0.0104     &     0.8247 $\pm$ 0.0032             &   0.7264 $\pm$ 0.0003        \\
		& GAT    &      0.8193 $\pm$ 0.0017            &         0.8556  $\pm$ 0.0009              &               0.7278 $\pm$ 0.0009    \\
		& Graph Transformer  &       \bf{0.8256} $\pm$ \bf{0.0031}              &    0.8560 $\pm$ 0.0003                   &   \bf{0.7311} $\pm$ \bf{ 0.0021 }   \\
		\multirow{-4}{*}{$\mathbf{X,A,\hat{Y}}$}  & \text{\ \ \ \ \ \  \ \ \ \ }$\llcorner$ w/ Edge Feature                      & *  &    \bf{0.8642} $\pm$ \bf{0.0008}    &  * \\
		\hline
	\end{tabular}
	}

	\caption{This is the ablation studies on models with different inputs, where
$\mathbf{X}$ denotes the nodes features, $\mathbf{A}$ is the graph adjacent matrix and $\mathbf{\hat{Y}}$ is the observed labels. In \emph{ogbn-proteins}, nodes features are not provided initially. We average the edge features as their nodes features and provide the experimental result of Transformer without edge features for fair comparison in this experiment, which is slightly different from Table \ref{table:ret_proteins}.
}
 \label{table:ablation-stuidies}
 \vspace{-4mm}
\end{table*}
 
 \subsection{Ablation Studies}
 In this section, to better identify the improvements from different components of our proposed model, we conduct extensive studies with the following four aspects:
 

\begin{itemize}
 	\item Firstly, we apply the masked label prediction strategy on kinds of GNNS to show the effectiveness and robustness of incorporation LPA and GNN, shown in Table \ref{table:ablation-stuidies}.
 	\item In order to get a more practical and effective solution to apply masked label prediction strategy, we tune the label\_rate during training and inference to explore the relationship between label coverage and GNNs performance, shown in Figure \ref{figure:fig_curves}.
 	\item We also analyze how LPA affects the GNN to make it performs better, shown in Figure \ref{fig:attenion_image}.
 	\item Furthermore, in Table \ref{table:ablation in unimp_v1}, we provide more ablation studies on UniMP, compared with GAT, showing the superiority of our model. 
\end{itemize}

 \subsubsection{Graph Neural Networks with Different Inputs}
 In Table \ref{table:ablation-stuidies}, we apply masked label prediction on kinds of GNNs to improve their performance. Firstly, we re-implement classical GNN methods like GCN and GAT, following the same sampling methods and model setting shown in Table \ref{table:params}. The hidden size of GCN is head\_num*hidden\_size since it doesn't have head attention. Secondly, we change different inputs for these models to study the effectiveness of feature and label propagation, using our {\bf masked label prediction} to train the models with partial nodes label $\mathbf{\hat{Y}}$ as input.
  
 
Row 4 in Table \ref{table:ablation-stuidies} shows that only with $\mathbf{\hat{Y}}$ and  $\mathbf{A}$ as input, GNNs still work well in all three datasets, outperforming those MLP model only given $\mathbf{X}$. This implies that one's label relies heavily on its neighborhood instead of its feature. Comparing Row 3 and 5 in Table \ref{table:ablation-stuidies}, models with $\mathbf{X}$, $\mathbf{A}$ and $\mathbf{\hat{Y}}$ outperform the models with $\mathbf{X}$ and $\mathbf{A}$, which indicates that it's a waste of information for GNNs in semi-supervised classification when they making predictions without incorporating the ground truth train labels $\mathbf{\hat{Y}}$. Row 3-5 in Table \ref{table:ablation-stuidies} also show that our Graph Transformer can outperform GAT, GCN with different input settings.

\subsubsection{Relation between Label Coverage and Performance}

 Although we have verified the effectiveness of using this strategy to combine LPA and GNN, the relation between label coverage and its impact on GNNs performance remains uncertain. Therefore, shown in Figure \ref{figure:fig_curves}, we conduct more experiments in \emph{ogbn-arxiv} to investigate their relationship in the following different scenarios:
\begin{itemize}
	\item In Figure \ref{figure:fig1}, we train UniMP using $\mathbf{X, \hat{Y}, A}$ as inputs. We tune the input label\_rate which is the hyper-parameter of masked label prediction task and display the validation and test accuracy. Our model achieves better performance when label\_rate is about 0.625.
	\item Figure \ref{figure:fig2} describes the correlation between the proportion of training data and the effectiveness of label propagation. We fix the input label\_rate with 0.625. The only change is the training data proportion. It's common sense that with the increased amount of training data, the performance is gradually improving. And the model with label propagation ${\mathbf{{\hat{Y}}}}$ can gain greater benefits from increasing labeled data proportion.
	\item Our unified model always masks a part of the training input label and tries to recover them. But in the inference stage, our model utilizes all training labels for predictions, which is slightly inconsistent with the one in training. In Figure \ref{figure:fig3}, we fix our input label\_rate with 0.625 during training and perform different input label\_rate in inference. the training stage, It's found that UniMP might have worse performance (less than 0.70) than the baseline (about 0.72) when lowering the label\_rate during prediction. However, when the label\_rate climbs up, the performance can boost up to 0.73.
	
	\item In Figure \ref{figure:fig4}, we calculate the accuracy for unlabeled nodes grouped by the number of neighbors. The experimental result shows that nodes with more neighbors have higher accuracy. And the model with label propagation ${\mathbf{\hat{Y}}}$ can always have improvements even with different numbers of training neighbors.
\end{itemize}

\subsubsection{Measuring the Connection between Nodes}

\begin{figure}[htpb]
\centering
	\begin{subfigure}[b]{.48\columnwidth}
		\includegraphics[width=\linewidth,trim= 0 0.2cm 0 0.2cm,clip]{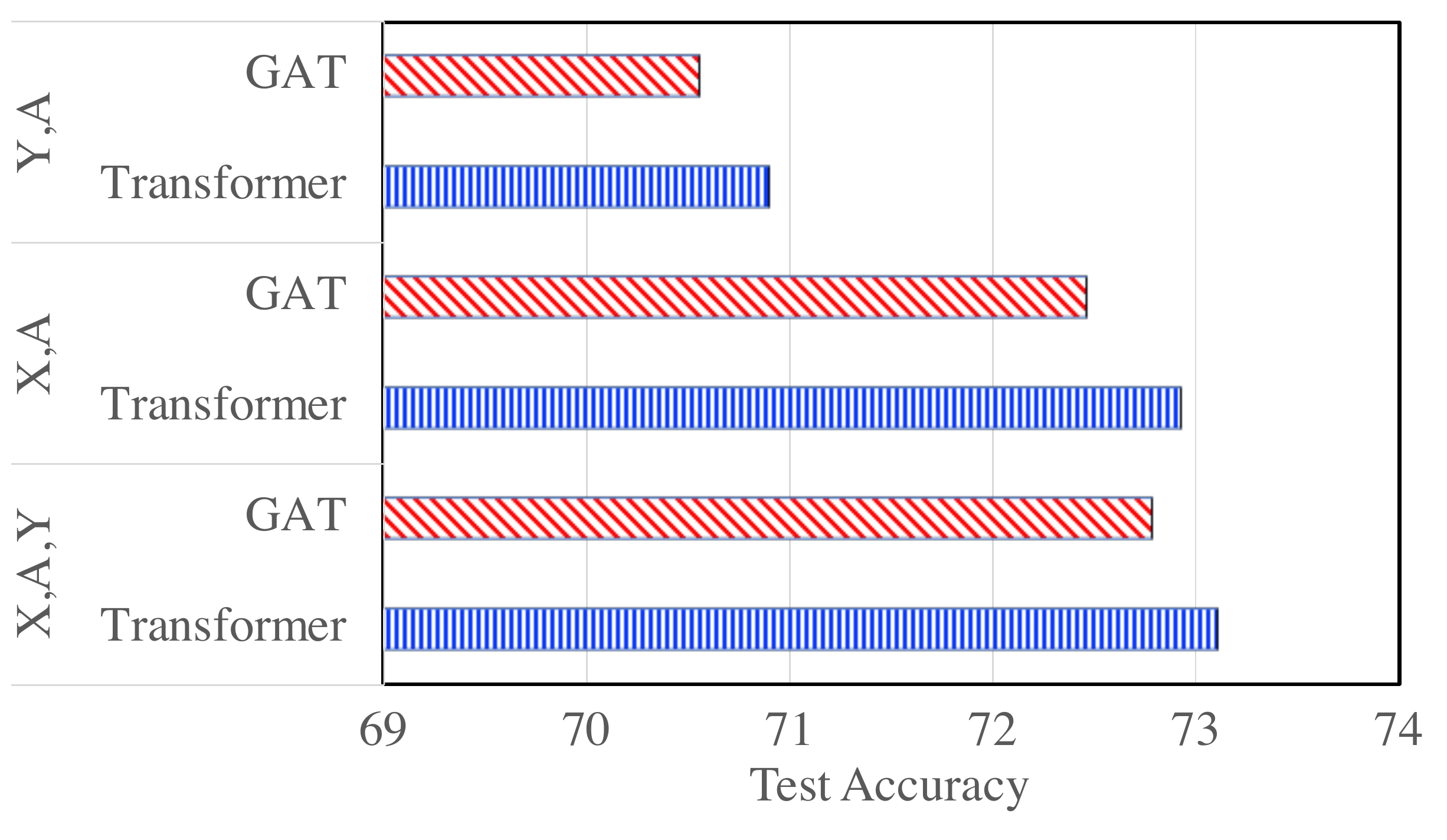}
		\label{figure:attention_accuracy}
	\end{subfigure}
	\begin{subfigure}[b]{.48\columnwidth}
		\includegraphics[width=\linewidth,trim= 0 0.2cm 0 0.2cm,clip]{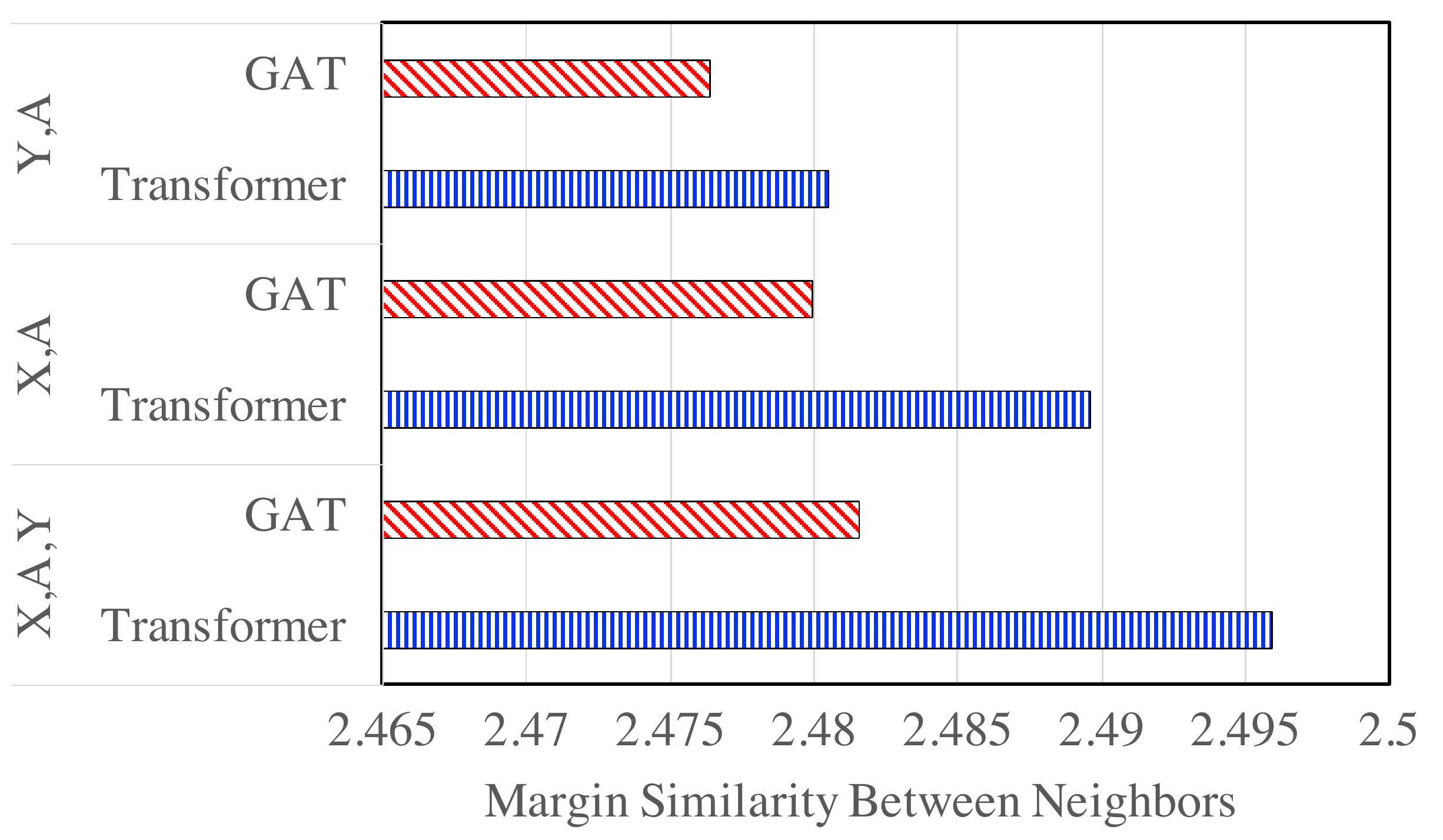}
		\label{figure:attention_margin}
	\end{subfigure}
	\setlength{\belowcaptionskip}{-.4cm}
	\caption{Correlation between accuracy and margin similarity between neighbors.}
	\label{fig:attenion_image}
\end{figure}

In Figure \ref{fig:attenion_image}, we analyze how LPA affects GNN to make it perform better. Wang~\shortcite{wang2019unifying} has pointed out that using LPA for GCN during training can enable nodes within the same class/label to connect more strongly, increasing the accuracy (ACC) of model's prediction. Our model can be regarded as an upgraded version of them, using LPA in both training and testing time for our Graph Transformer. Therefore, we try to experimentally verify the above idea based on our model.

\begin{equation}
\resizebox{.80\linewidth}{!}{$
\begin{aligned}
MSF =\dfrac{1}{N}\sum_{i=1}^{N} \log \bigg ( 1+ \sum_{j\in \mathcal{N}(i)_{pos}} \sum_{k\in \mathcal{N}(i)_{neg}}e^{\alpha_{i,j}}-e^{\alpha_{i,k}} \bigg)\\
\end{aligned}
\label{eq:11}$}
\end{equation}

We use the Margin Similarity Function (MSF) as shown in Equation \ref{eq:11} to reflect the connection tightness between nodes within the same class (the higher scores, the stronger connection they have. We conduct the experiment on \emph{ogbn-arxiv}. And as shown in Figure \ref{fig:attenion_image}, the ACC of models' prediction is proportional to Margin Similarity. Unifying feature and label propagation can further strengthen their connection, improving their ACC. Moreover, our Graph Transformer outperforms GAT in both connection tightness and ACC with different inputs.

\subsubsection{More Ablation Studies on UniMP}

Finally, we provide more ablation studies on our UniMP model, compared with GAT, from the following 4 aspects: 
(1) vanilla transformer with dot-product attention or GAT with sum attention; 
(2) simple residual or gated residual;
(3) with train labels as inputs;
(4) with train and validation labels as inputs.
As shown in Table \ref{table:ablation in unimp_v1}, we can find that dot-product attention can outperform sum attention, since dot-product provides more interactions between nodes. Besides, residual and gated residual can also strengthen the GNNs with shallow layers. Moreover, our unified model can take the additional validation labels as input to further boost model's performance without more training steps. Therefore, when we apply the model to the real scene, and the labeled data are accumulated progressively, the accuracy of the unlabeled data can keep increasing without training our model from scratch, while other GNNs without explicit label modeling can't fully utilize the benefits of additional labels.

\begin{table}[htbp]
\resizebox{\columnwidth}{!}{
\begin{tabular}{ccc}
\hline
 Model                                                                                                               & \textbf{ogbn-prdouct}                                                     & \textbf{ogbn-arxiv}                                                       \\ \hline
\begin{tabular}[l]{@{}l@{}}GAT (sum attention)\\ $\llcorner$ w/ residual \\ $\llcorner$ w/ gated residual\end{tabular}                                          & \begin{tabular}[c]{@{}c@{}} 0.8002 \\ 0.8033\\ 0.8050\end{tabular}                   & \begin{tabular}[c]{@{}c@{}} 0.7246 \\ 0.7265\\ 0.7272\end{tabular}                   \\ \hline
\begin{tabular}[l]{@{}l@{}}Transformer (dot-product)\\ $\llcorner$ w/ residual \\ $\llcorner$ w/ gated residual\\ ~~~~$\llcorner$ w/ train label (UniMP) \\  ~~~~~~~~$\llcorner$ w/ validation labels\end{tabular} & \begin{tabular}[c]{@{}c@{}} 0.8091\\ 0.8125\\ 0.8137\\ 0.8256\\ \bf{0.8312} \end{tabular} & \begin{tabular}[c]{@{}c@{}} 0.7259 \\ 0.7271 \\ 0.7292\\ 0.7311\\ \bf{0.7377}\end{tabular} \\ \hline
\end{tabular}}
\caption{Ablation studies in UniMP, compared with GAT}
	\label{table:ablation in unimp_v1}
	\vspace{-4mm}
\end{table}

	\section{Conclusion}
	We first propose a unified message passing model, UniMP, which jointly performs feature propagation and label propagation within a Graph Transformer to make the semi-supervised classification. Furthermore, we propose a masked label prediction method to supervised training our model, preventing it from overfitting in self-loop label information. Experimental results show that UniMP outperforms the previous state-of-the-art models on three main OGBN datasets: \emph{ogbn-products}, \emph{ogbn-proteins} and \emph{ogbn-arxiv} by a large margin, and ablation studies demonstrate the effectiveness of unifying feature propagation and label propagation.
	
	\bibliographystyle{named}
	\bibliography{ijcai21}
	
\end{document}